\documentclass[sigconf]{acmart}
\AtBeginDocument{%
  \providecommand\BibTeX{{%
    \normalfont B\kern-0.5em{\scshape i\kern-0.25em b}\kern-0.8em\TeX}}}

\setcopyright{acmcopyright}
\copyrightyear{2025}
\acmYear{2025}
\acmDOI{XXXXXXX.XXXXXXX}
\usepackage{graphicx}
\usepackage{amsmath,amsfonts}
\usepackage{float}
\usepackage{textcomp}
\usepackage{xcolor}
\usepackage{times}
\usepackage{epsfig}
\usepackage{graphicx}
\usepackage{amsmath}
\usepackage{makecell}
\usepackage{tabularx,booktabs} 
\usepackage{amsmath}
\usepackage{mathtools}
\usepackage{multirow}
\usepackage{algorithm}
\usepackage{algpseudocode}
\usepackage{balance}
\usepackage{subcaption}
\usepackage{enumitem}
\usepackage{paralist}
\usepackage{xcolor}
\usepackage{hyperref}
\hypersetup{colorlinks=true,linkcolor=blue,urlcolor=blue}
\acmConference[ACM IKDD CODS 2025]{the 13th International Conference on Data Science}{December 17--20, 2025}{Pune, India}
\begin{document}

\title{Robust Face Liveness Detection for Biometric Authentication using Single Image}

\author{Poulami Raha}
\affiliation{%
  \institution{Rakuten Group Inc.}
  \country{}
  \city{}}
\email{poulami.raha@rakuten.com}
\author{Yeongnam Chae}
\affiliation{%
  \institution{Rakuten Group Inc.}
  \country{}
  \city{}}
\email{yeongnam.chae@rakuten.com}

\renewcommand{\shortauthors}{Raha et al.}

\begin{abstract}
Biometric technologies are widely adopted in security, legal, and financial systems. Face recognition can authenticate a person based on the unique facial features such as shape and texture. However, recent works have demonstrated the vulnerability of Face Recognition Systems (FRS) towards presentation attacks. Using spoofing (aka., presentation attacks), a malicious actor can get illegitimate access to secure systems. This paper proposes a novel light-weight CNN framework to identify print/display, video and wrap attacks. The proposed robust architecture provides seamless liveness detection ensuring faster biometric authentication (1-2 seconds on CPU). Further, this also presents a newly created 2D spoof attack dataset consisting of more than 500 videos collected from 60 subjects. To validate the effectiveness of this architecture, we provide a demonstration video depicting print/display, video and wrap attack detection approaches. The demo can be viewed in the following link:\\
\url{https://rak.box.com/s/m1uf31fn5amtjp4mkgf1huh4ykfeibaa}

\end{abstract}

%



\maketitle

\section{Introduction}\label{sec1}

Biometric authentication has been recently preferred either over or in addition to traditional approaches due to increased security and hassle-free user convenience. Of the many available biometric characteristics such as iris, fingerprint, vein and face characteristics, face is preferred in many applications such as identity management \cite{jain2004introduction,jain2006biometrics}, online payment \cite{chen2005cashless}, access control \cite{jain2006biometrics, jain2011handbook}, automotive applications and active authentication on mobile \cite{fathy2015face, mahbub2016active}. Face Recognition Systems (FRS) are susceptible to various attacks at various points within the operational system, starting from capture level to comparison sub-system \cite{jain2012biometric}. Among all types of attacks, such as make-up attacks \cite{bharati2017demography, rathgeb2019impact}, mask-attacks \cite{ agarwal2017face, marcel2014handbook, liu2018detecting}, spoofing attacks through printed, display and replay (image/video) attacks \cite{marcel2014handbook, erdogmus2013spoofing, de2012lbp, anjos2014face}, the presentation attacks are reported to pose severe threats \cite{agarwal2017face, mehta2019crafting, agarwal2019deceiving, majumdar2019evading}. 

\par While the advanced attacks using sophisticated 3D masks created using Latex material or Silicone are challenging to detect, they are also very expensive to conduct \cite{vareto2019swax, bhattacharjee2019recent, kotwal2019multispectral}. Several anti-spoofing techniques have been proposed for efficient protection against various sophisticated spoofing attacks such as print/display, video attacks or expensive silicone mask attacks \cite{csmask, kotwal2019multispectral}. The simplistic attacks using printed photo or display attacks are easy to conduct, they threaten the FRS equally well \cite{ komulainen2019review, ramachandra2017presentation, mehta2019crafting, agarwal2019deceiving, majumdar2019evading}. 

\par In 2D spoof attacks, a malicious actor uses a printed photo, display devices showing the photo, or a looped video of an authentic user's face. Print attacks are the most common type of attack that can be conducted using cut photo and warped photo attacks \cite{hernandez2019introduction, komulainen2019review, evans2019handbook, boulkenafet2018generalization}. 

In this work, we propose a novel light-weight CNN framework to reliably detect 2D print/display, video and wrap attacks without relying on advanced imaging methodologies such as spectral imaging or RGB-D imaging. Our proposed network is extremely light-weight in nature, the network is carefully designed to have less parameters, which ensures faster authentication (1-2 seconds on CPU). While in many works, the approaches to detect the spoofing attacks are based on complex and large architectures, we propose a robust face liveness detection model for various spoof attacks, which along with usual facial features also utilizes face image background information effectively to discriminate between the live user and spoof attack. Usually, live user face will appear with normal background, whereas attack samples will appear with paper or screen background. The proposed novel architecture trains well in our finely distributed dataset, and effectively discriminates between the real and attack samples, similarly between live user background and paper or screen background. This strategy ensures accurate prediction of various kind of spoof attacks. The following are the primary contributions:

\vspace{0.1cm}
\begin{compactitem} 
\item A novel light-weight CNN framework to reliably detect print or display, video and wrap attacks. The proposed architecture provides seamless liveness detection for biometric authentication (1-2 seconds on CPU).

\item 2D spoof attack dataset of 60 subjects with over 500 videos of  print/display, video and wrap attacks of different conditions.

\item Demonstration video depicting print/display, video and wrap attack detection approaches using a web application. 
\end{compactitem}
\vspace{0.1cm}

The paper is organized as follows: first, we provide a brief review of related works in Section \ref{sec2}. In Section \ref{sec3}, the proposed methodology, network architecture and an overview of the newly collected dataset are discussed. In Section \ref{sec4}, we present demonstration details and analysis, and finally, Section \ref{sec5} concludes the paper.

\section{Related Work}\label{sec2}
In this section, we present a brief overview of recent deep learning-based feature-level face-anti spoofing methodologies or presentation attack detection (PAD) algorithms. 

\par Anti-spoofing methodologies can be categorized into three main groups, namely sensor-level (hardware-based), feature-level(software-based), and score-level (the combination of sensor-level and feature-level techniques). In \cite{3dmad}, Erdogmus {\textit{et al.}} have provided a 3D facial mask dataset along with LBP texture analysis against the mask based face spoofing attacks. Their experimentation shows that the LDA provides better results compared to SVM with HTER of 0.95\% and 1.27\% for color and depth images respectively on 3D MAD dataset.

\par Manjani {\textit{et al.}} \cite{ddgl} uses a multi-level deep dictionary via a greedy learning algorithm to prevent silicone mask based spoof attacks. The proposed approach with SVM classifier reports HTER of 0.0\% on 3DMAD and Replay-Attack, 1.3\% , 16.5\%, and 13.1\% on CASIA-FASD, UVAD, and SMAD datasets. Li {\textit{et al.}} \cite{rose} introduced an unsupervised domain adaptation framework using both hand-crafted and deep neural network learned features. In this framework, Local Phase Quantization (LPQ) and Co-occurrence of Adjacent Local Binary Patterns (CoALBP) with Convolutional Neural Network (CNN) based texture features are extracted. The approach mentioned above improves the overall generalization capability of an anti-spoofing system with an improvement of an average of 20\% when evaluated in Replay-Attack, ROSE-Youtube, CASIA, and MSU FASD datasets. 

\par Bhattacharjee {\textit{et al.}} \cite{csmask} has presented an intensity over thermal imagery-based anti-spoofing countermeasures with a detailed analysis of state-of-the-art CNN based face recognition systems. VGG-Face, Light CNN, and FaceNet network report the highest vulnerability with IAPMR of 57.9\%, 22.6\%, and 42.1\% respectively on the CSMAD dataset. Ramachandra {\textit{et al.}} \cite{psan} have presented an empirical survey of spoofing attacks and the venerability of two different commercial face recognition systems against custom silicone masks. 

\par Liu {\textit{et al.}} \cite{rppg} has presented deep learning-based CNN-RNN architecture to estimate remote photoplethysmography (rPPG) signals from face videos to mitigate the spoofing attempts. CNN part provides the face depth map estimation with the help of pixel-based depth loss function. The Recurrent Neural Network (RNN) utilizes chromium signals and FFT to measure rPPG signals. The proposed method provides cross-testing HTER of 27.6\% and 28.4\% for CASIA-MFSD and Replay-Attack datasets, respectively. Rehman {\textit{et al.}} \cite{livenet} have provided a CNN (VGG-11) based methodology with the help of continuous data-randomization and  mini-batches training. This methodology has reported HTER of 8.28\% and 14.14\% using cross-dataset testing on CASIA-FASD and Replay-Attack dataset, respectively. Several other CNN-based approaches have been reported in the following papers \cite{S1,deep2}. 

\section{Methodology}\label{sec3}
In this section, We briefly go over the few aspects of the proposed liveness detection methodology:
\\
\textbf{Development of robust light-weighted CNN model:}\\
The proposed CNN is thoughtfully designed to be light-weighted with few parameters, total number of parameter reported is around 170k, which makes the model extremely robust. The details of the proposed architecture is given in Table \ref{tab:table0}. The architecture is robust in nature, which ensures seamless face liveness detection in many applications. The traditional state-of-the-art architectures are complex, these kind of networks are hard to train, extremely computationally expensive and heavy in size. The high performance light models are much faster, easier to infer, computationally inexpensive and useful for many applications such as mobile or web applications. The Lightweight CNN models always offer better performance as compared to heavy-weight models, targeting real-time applications. 
\\
\textbf{Create balanced dataset for spoof attack detection:}\\
The end-to-end training of the proposed network is done using our in-house datasets. We utilize few strategies to curate our training dataset. First, We create a well balanced dataset of standard cropped face images of bonafide and attack classes, the attack class consists of print/display, video and wrap attack samples which includes several distance and lighting conditions. Second, We also include many cropped face images with extra padding to ensure inclusion of user background information along with face features. In case of live user background will be normal and attack samples will contain paper or screen background. A brief description of our dataset is given in the next section.
\par The proposed novel architecture trains well in our finely distributed dataset, and effectively discriminates between the real and attack samples, similarly between live user background and paper or
screen background, while preserving the robustness of the model. The proposed approach achieves best ACER of 2.7\% in our newly created dataset.

\begin{table*}[h]
\caption{Proposed Liveness Model Architecture}\label{tab:table0}
 \begin{tabular}{|c|c|c|}
 \hline
  \makecell{
  3x3 16 Conv, BNorm, ReLU \\ 
  3x3 16 Conv, BNorm, ReLU \\ 
  3x3 16 Conv, BNorm, ReLU \\ 
  3x3 16 Conv, BNorm, ReLU \\
  2x2 Max Pool, Dropout \\
} & \makecell{
  3x3 32 Conv, BNorm, ReLU \\ 
  3x3 32 Conv, BNorm, ReLU \\ 
  3x3 32 Conv, BNorm, ReLU \\ 
  3x3 32 Conv, BNorm, ReLU \\
  2x2 Max Pool, Dropout \\
} & \makecell{
  Latent Vector Representation \\
  Fully-connected layer, ReLU \\
  BNorm, Dropout \\
  Fully-connected layer, Softmax\\
  \textit{Total Parameters: 170k}} \\ 
 \hline
\end{tabular}
\end{table*}

\subsection{Spoof Attack Dataset}
In order to address 2D spoof attacks, we created a new face spoof dataset. Our dataset consists of more than 500 videos where bona fide and spoof attack videos are captured from 60 subjects. The acquisition device used for data collection is iPad Pro 11'' with a tripod stand for better stability. The data is captured indoors in normal lighting conditions and at two different distances between subjects and acquisition device. 

\subsubsection{Bona fide/Real Data}

\begin{figure}[h]
	\hspace{1.5em}\includegraphics[width=0.45\textwidth]{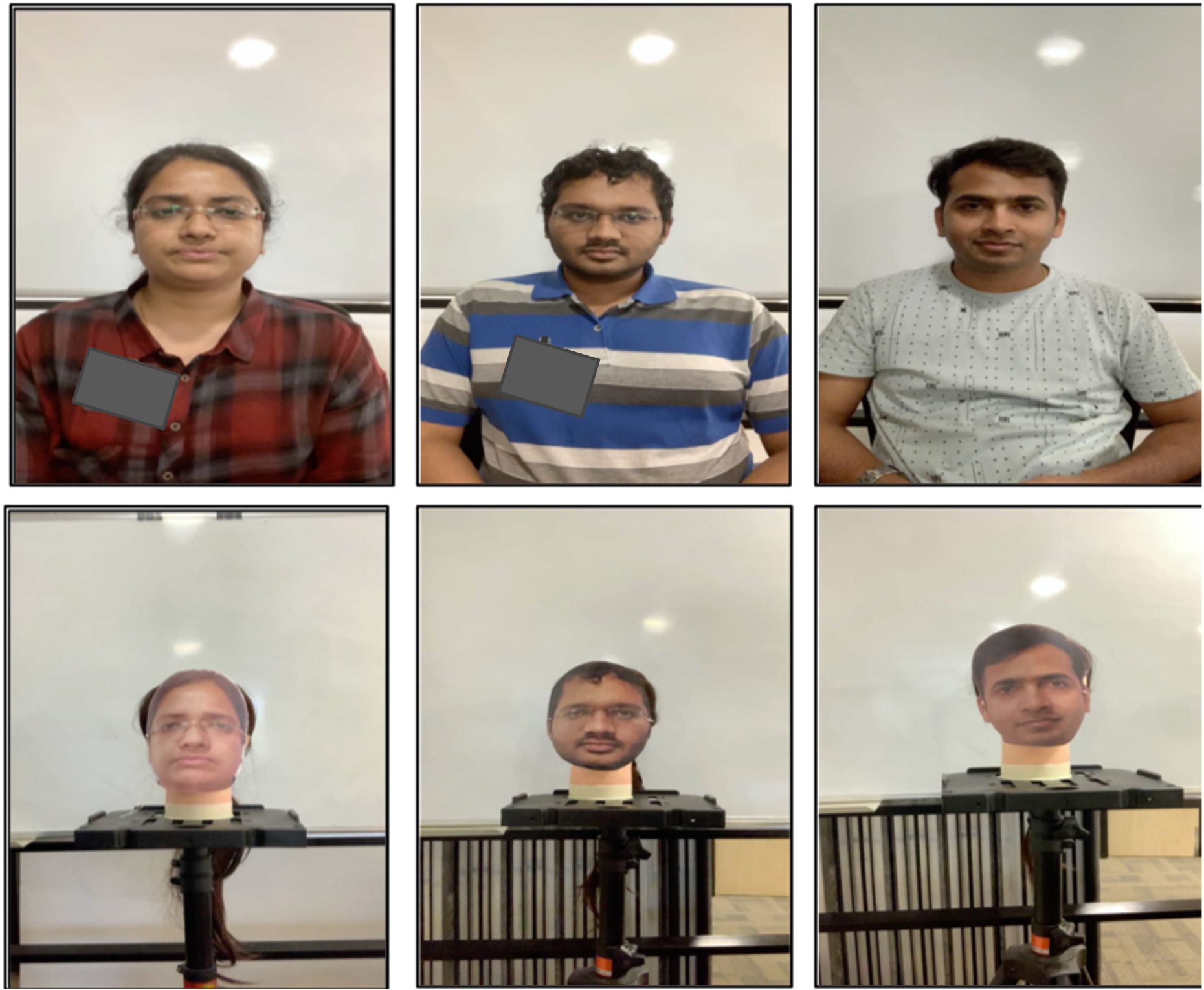}
	\caption{Experimental Setup.}
    \Description{}
        \label{fig:Experimental}
\end{figure}

\begin{figure}[h]
	\centering	
	\begin{subfigure}{0.5\textwidth}
		\centering
		\includegraphics[width=0.8\linewidth]{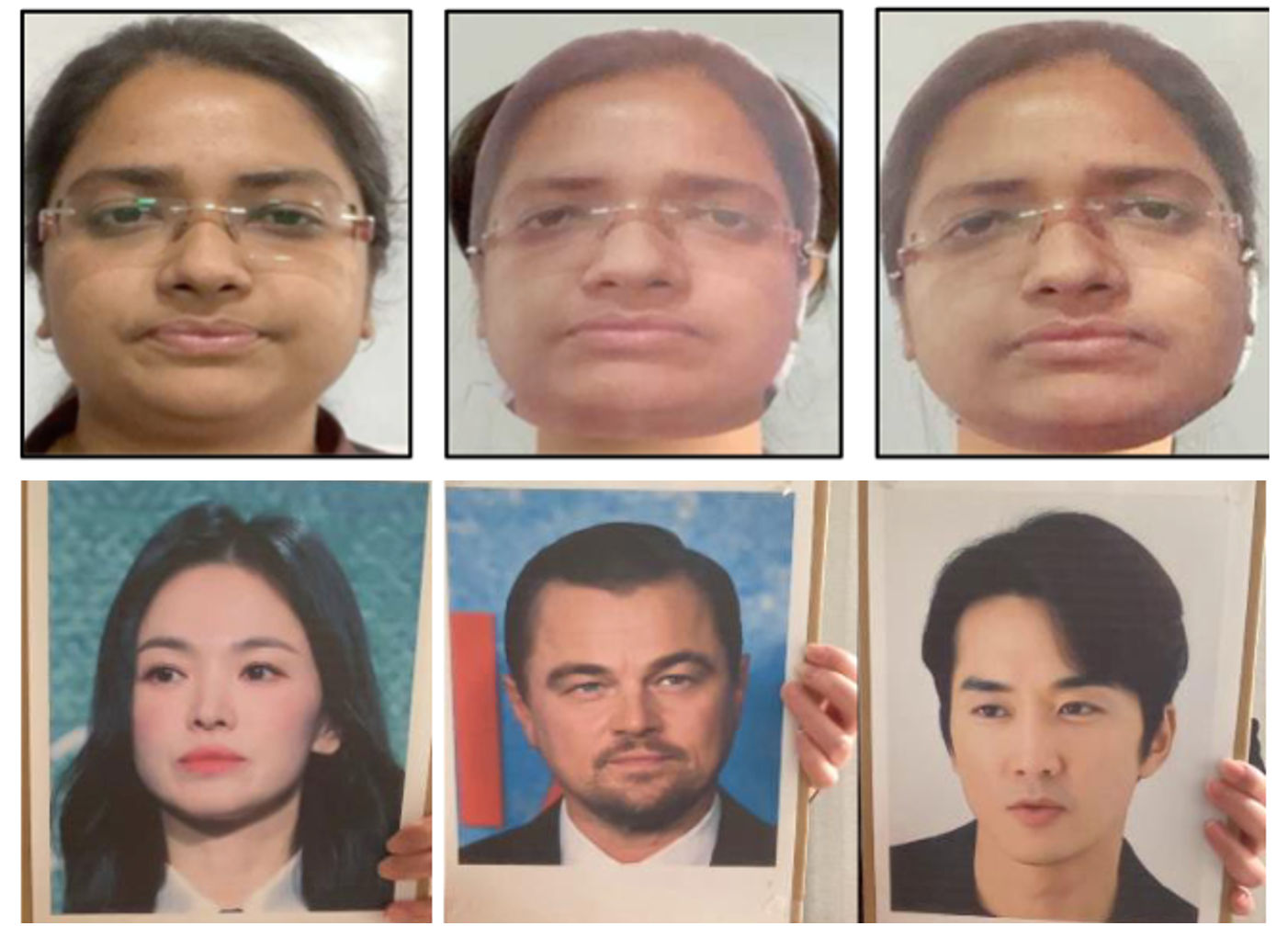}
	\end{subfigure}
 	\begin{subfigure}{0.5\textwidth}
		\centering
		\includegraphics[width=0.8\linewidth]{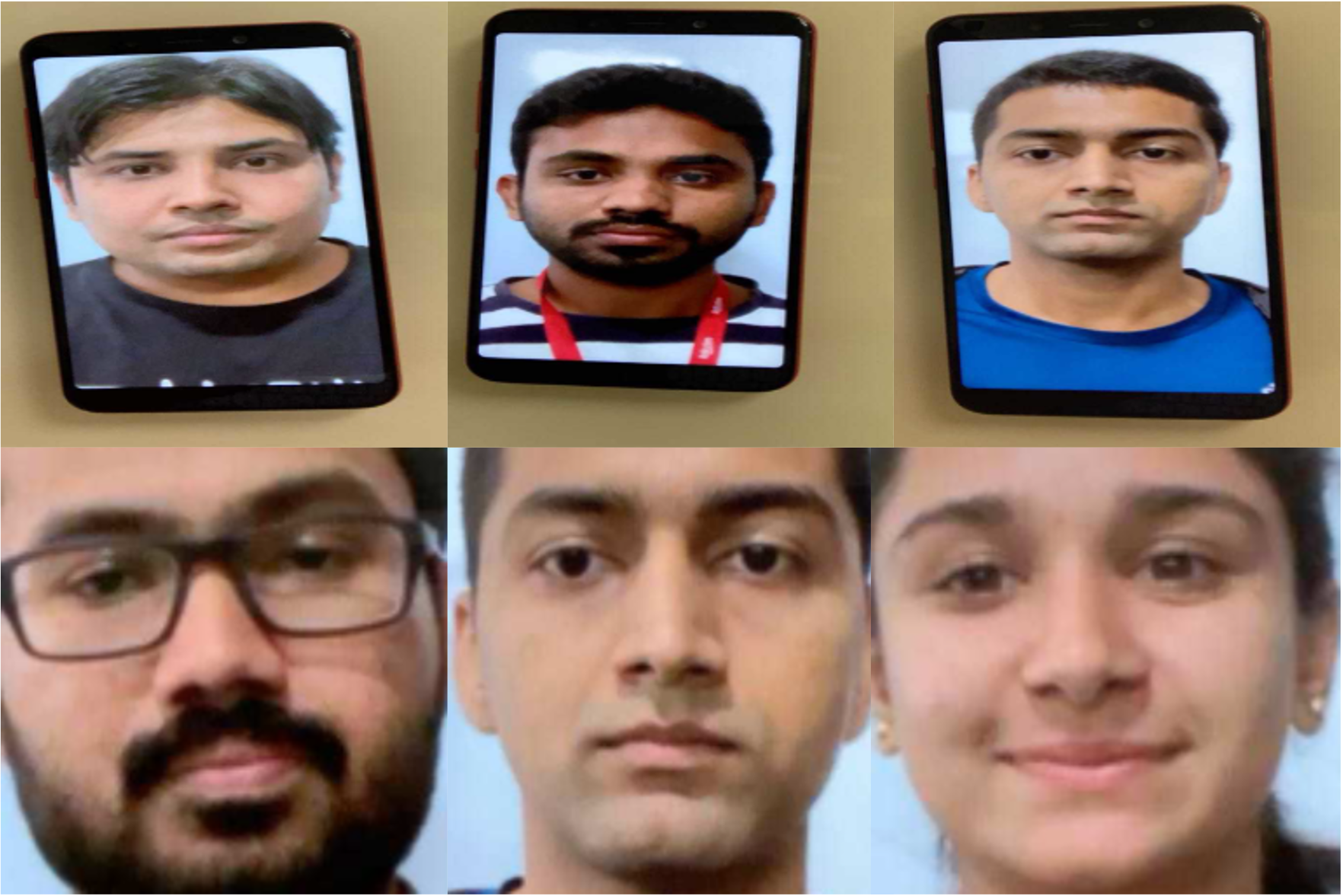}
	\end{subfigure}
	\caption{Cropped Face Images with/without Padding.}
    \Description{}
 	\label{fig:Sample}
\end{figure}

In total, there are 60 subjects data in our dataset. As mentioned earlier, two different types of bonafide videos are captured based on the distance between subjects and acquisition device, namely Mid and Close distance. The stand-off distances between the subjects and the camera for Mid and Close type were $35-37$ inches and $8-9$ inches for respectively. In total, 60 subjects were asked to provide their data for mid-distance scenario, whereas for the close distance, total 14 subjects are asked to provide their data again. Each bona fide samples were recorded for 10 seconds. Fig. \ref{fig:Experimental} and Fig. \ref{fig:Sample} shows the experimental setup and sample images from our dataset.

\subsubsection{Spoof/Attack Data}
Total five types of attack videos of 60 subjects are generated with mid and close distance variations:
\begin{compactitem} 
\item Normal Color Print (Print photo paper) 
\item Color Print (Glossy photo paper) 
\item Video (Mobile/iPad screen)  
\item Normal Color Printed Mask (Print photo paper)
\item Color Printed Mask (Glossy photo paper) 
\end{compactitem} 
A mannequin head and rigid stand are used to capture the wrap attack videos. The distance between different displays and the camera were $35-37$ inches and $23-25$ inches, respectively. The captured attack videos are of length 10 seconds. Fig. \ref{fig:Experimental} and Fig. \ref{fig:Sample} shows the experimental setup and few sample images from our dataset.

\subsubsection{Training Details}
The proposed methodology utilizes the MTCNN network for data pre-processing such as face and landmark detection, scaling, cropping of facial regions with or without padding \cite{mtcnn}. The extra padding of cropped faces provides us with user background information, the live user face will appear normal, whereas in case of attack user face will appear with paper or screen background. This strategy ensures accurate prediction of various kind of spoof attacks. Normalization is applied to the face images to restrict the dynamic range of the input RGB images to [0,255]. We have used around 3.5K face images per class for real and attack samples, which includes various print/display/video/wrap attack samples. The input size of the network is set to 32x32x3, with a batch size of 32. In the features extraction stage, the features were extracted from the convolution network's last flattened layer. Finally, fully-connected layers provides the prediction based on user face images which along with usual face images includes background information. 

\par The statistical information of the proposed dataset is given in Table \ref{tab:data}. Based on subjects, the proposed dataset is divided into 3 subsets: training, development and testing. 

\begin{table}[h]
\caption{Statistical information of the proposed dataset}\label{tab:data}
\begin{tabular}{|c|c|c|}
\hline
dataset     & Subjects & Images  \\
\hline
Training     & 30                 & 7000  \\
\hline
Development & 20                 & 2000  \\
\hline
Testing     & 10                 & 1000  \\
\hline
\end{tabular}
\end{table}

\begin{figure*}[h]
	\centering
	\hspace{1.5em}\includegraphics[width=0.9\textwidth]{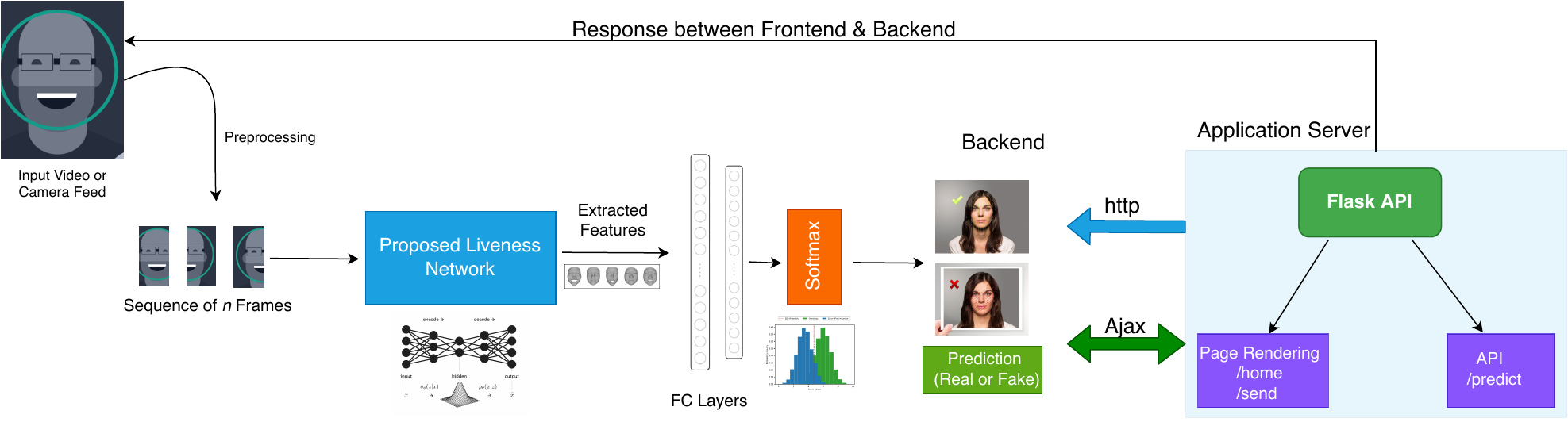}
	\caption{Schematic diagram of web demonstrator for detecting face spoof attacks.}
    \Description{}
	\label{fig:web}
\end{figure*}
 
\begin{figure}[h]
	\centering
	\hspace{1.5em}\includegraphics[width=0.5\textwidth]{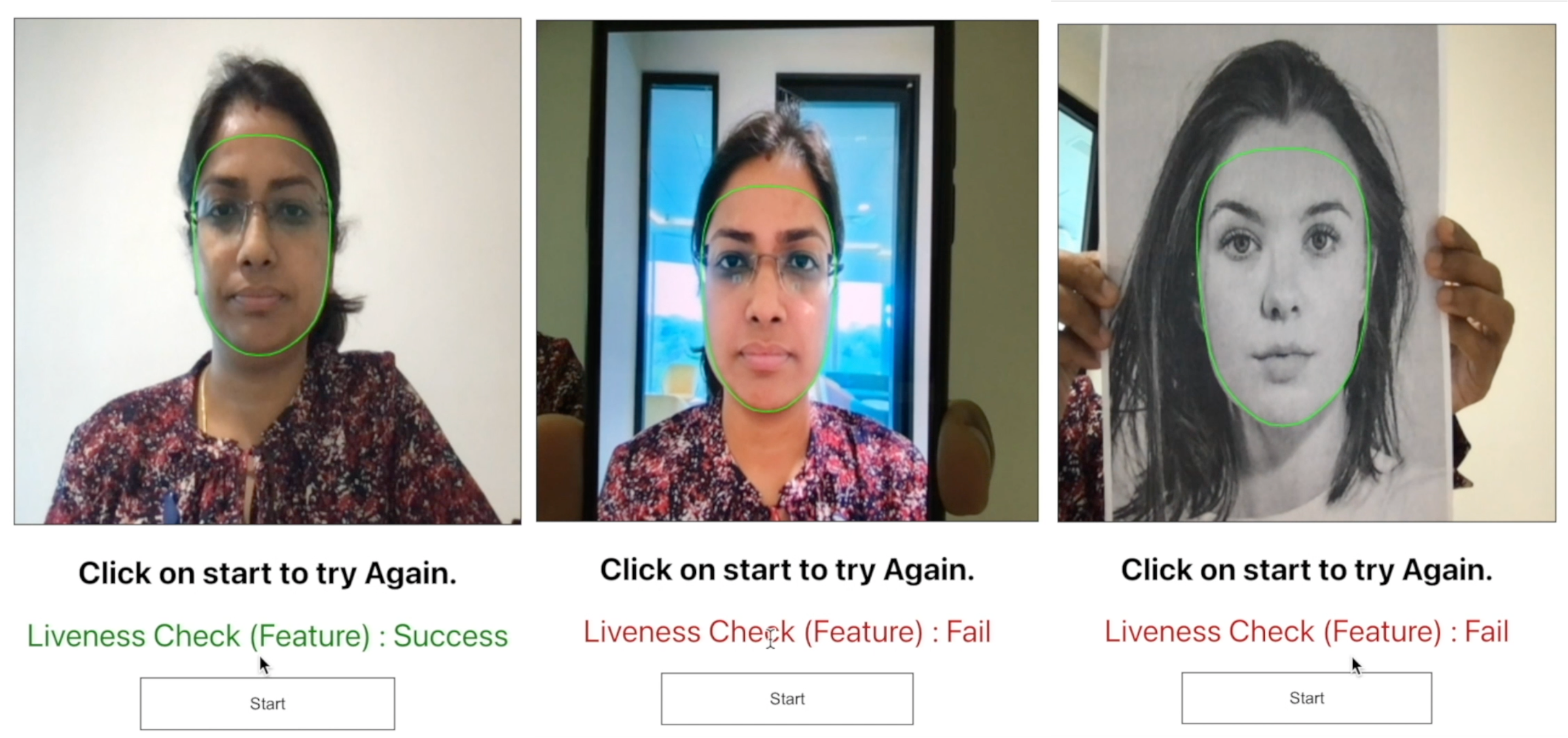}
	\caption{Screenshots of the web demonstrator (Display Attacks)}
    \Description{}
	\label{fig:demo1}
\end{figure}

\begin{figure}[h]
	\centering
	\hspace{1.5em}\includegraphics[width=0.5\textwidth]{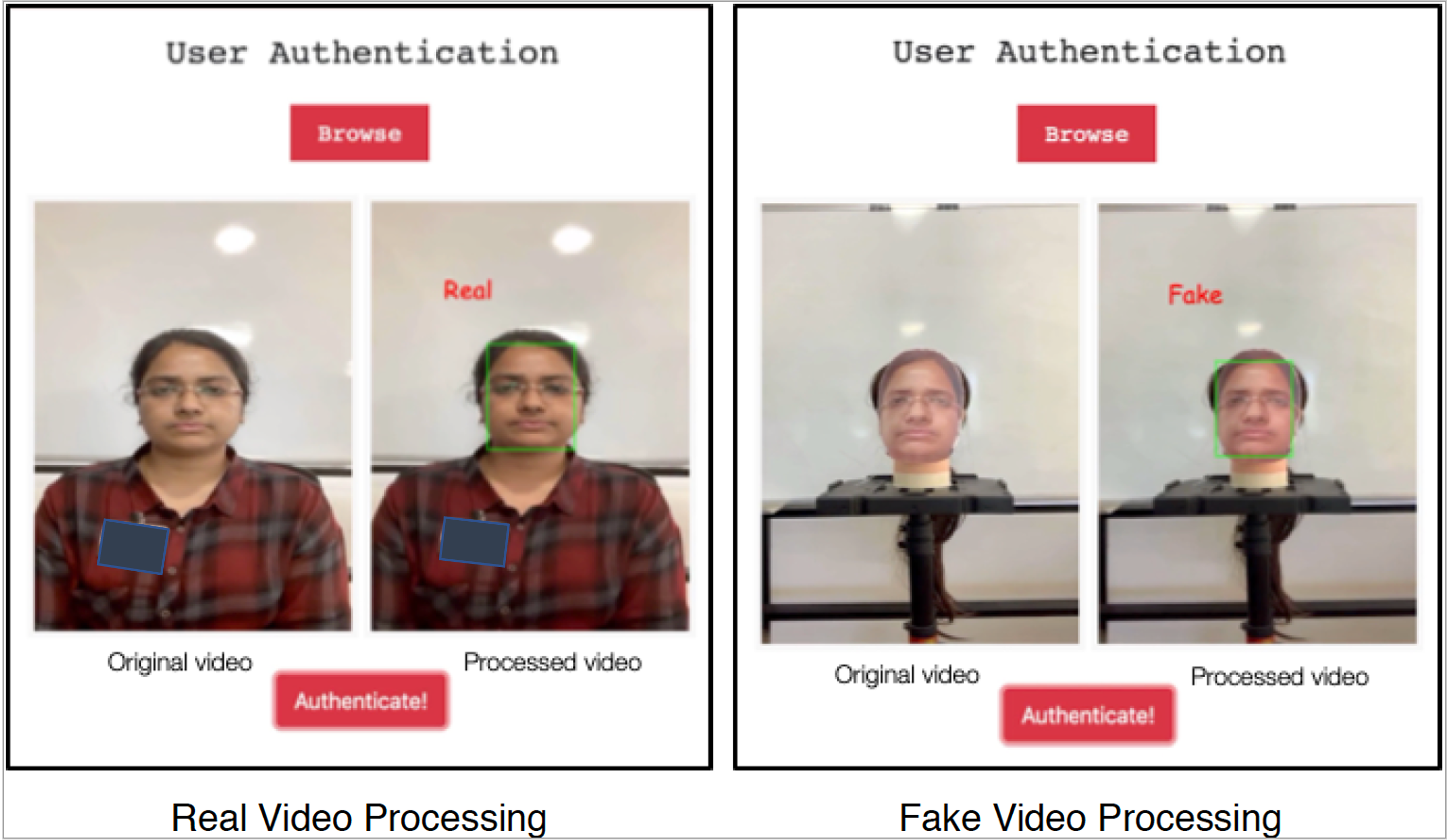}
	\caption{Screenshots of the web demonstrator (Wrap Attack)}
    \Description{}
	\label{fig:demo2}
\end{figure}

\section{Demonstration Interface}\label{sec4}
In order to comprehend the real-time analysis and effectiveness of the proposed methodology, we have developed a demonstrator in terms of a simple web application. Fig. \ref{fig:web} shows the workflow of the web application. Firstly, we need to process input video or camera feed frame by frame, i.e., a frame from the camera or video is captured and uploaded to the server for the processing. At the server-side, we first detect the face and facial landmarks using MTCNN \cite{mtcnn}, then the cropped face is forward passed to our proposed network, and features are extracted from the flattened layer. These features are then passed to the fully-connected layers for obtaining the inference. The obtained label and face bounding box details are then sent back to the client-side application for rendering the output. 

\par Fig. \ref{fig:demo1}, \ref{fig:demo2} presents the screenshots of the demonstrator depicting its effectiveness in seamless face liveness detection. Fig. \ref{fig:demo1} showcases real-time successful liveness check for the real faces, and detection of print/display and video attacks respectively. Similarly, \ref{fig:demo2} provides real-time liveness check for wrap attack detection. 

\par The primary objective of our web-demonstrator was to verify and showcase real-time applications of our robust liveness detection model which is fast, compact, and can provide seamless biometric authentication using minimal hardware resources. The video demonstrates the performance of our proposed lightweight CNN, which has a very high processing speed for face liveness detection, with a processing time of 1-2 seconds on CPU. \\
\textbf{The demo can be viewed in the following link:} \\
\url{https://rak.box.com/s/m1uf31fn5amtjp4mkgf1huh4ykfeibaa}

\section{Conclusion}\label{sec5}
In this paper, we have presented a light-weight CNN architecture for various spoof attack detection. We also present a newly created spoof attack dataset. A single end-to-end CNN is proposed to efficiently learn the latent features of bonafide and attack face images to discriminate between the bonafide and attack users. The proposed real-time liveness detection, which processes faces, takes 1-2 seconds at most on CPU. We provide a demonstration video seamlessly conducting liveness detection on various spoof attacks. Our web demonstration validates that the proposed light-weight CNN is capable of dealing with various sophisticated spoof attacks.


\begin{thebibliography}{36}


\ifx \showCODEN    \undefined \def \showCODEN     #1{\unskip}     \fi
\ifx \showDOI      \undefined \def \showDOI       #1{#1}\fi
\ifx \showISBNx    \undefined \def \showISBNx     #1{\unskip}     \fi
\ifx \showISBNxiii \undefined \def \showISBNxiii  #1{\unskip}     \fi
\ifx \showISSN     \undefined \def \showISSN      #1{\unskip}     \fi
\ifx \showLCCN     \undefined \def \showLCCN      #1{\unskip}     \fi
\ifx \shownote     \undefined \def \shownote      #1{#1}          \fi
\ifx \showarticletitle \undefined \def \showarticletitle #1{#1}   \fi
\ifx \showURL      \undefined \def \showURL       {\relax}        \fi
\providecommand\bibfield[2]{#2}
\providecommand\bibinfo[2]{#2}
\providecommand\natexlab[1]{#1}
\providecommand\showeprint[2][]{arXiv:#2}

\bibitem[Agarwal et~al\mbox{.}(2019)]%
        {agarwal2019deceiving}
\bibfield{author}{\bibinfo{person}{Akshay Agarwal}, \bibinfo{person}{Akarsha Sehwag}, \bibinfo{person}{Mayank Vatsa}, {and} \bibinfo{person}{Richa Singh}.} \bibinfo{year}{2019}\natexlab{}.
\newblock \showarticletitle{Deceiving the protector: Fooling face presentation attack detection algorithms}.
\newblock \bibinfo{journal}{\emph{IEEE/IAPR ICB}} \bibinfo{volume}{1}, \bibinfo{number}{5} (\bibinfo{year}{2019}).
\newblock


\bibitem[Agarwal et~al\mbox{.}(2017)]%
        {agarwal2017face}
\bibfield{author}{\bibinfo{person}{Akshay Agarwal}, \bibinfo{person}{Daksha Yadav}, \bibinfo{person}{Naman Kohli}, \bibinfo{person}{Richa Singh}, \bibinfo{person}{Mayank Vatsa}, {and} \bibinfo{person}{Afzel Noore}.} \bibinfo{year}{2017}\natexlab{}.
\newblock \showarticletitle{Face presentation attack with latex masks in multispectral videos}. In \bibinfo{booktitle}{\emph{Proceedings of the IEEE Conference on Computer Vision and Pattern Recognition Workshops}}. \bibinfo{publisher}{IEEE}, \bibinfo{address}{USA}, \bibinfo{pages}{81--89}.
\newblock


\bibitem[Anjos et~al\mbox{.}(2014)]%
        {anjos2014face}
\bibfield{author}{\bibinfo{person}{Andr{\'e} Anjos}, \bibinfo{person}{Jukka Komulainen}, \bibinfo{person}{S{\'e}bastien Marcel}, \bibinfo{person}{Abdenour Hadid}, {and} \bibinfo{person}{Matti Pietik{\"a}inen}.} \bibinfo{year}{2014}\natexlab{}.
\newblock \showarticletitle{Face anti-spoofing: Visual approach}.
\newblock In \bibinfo{booktitle}{\emph{Handbook of biometric anti-spoofing}}. \bibinfo{publisher}{Springer}, \bibinfo{address}{Germany}, \bibinfo{pages}{65--82}.
\newblock


\bibitem[Bharati et~al\mbox{.}(2017)]%
        {bharati2017demography}
\bibfield{author}{\bibinfo{person}{Aparna Bharati}, \bibinfo{person}{Mayank Vatsa}, \bibinfo{person}{Richa Singh}, \bibinfo{person}{Kevin~W Bowyer}, {and} \bibinfo{person}{Xin Tong}.} \bibinfo{year}{2017}\natexlab{}.
\newblock \showarticletitle{Demography-based facial retouching detection using subclass supervised sparse autoencoder}. In \bibinfo{booktitle}{\emph{2017 IEEE International Joint Conference on Biometrics (IJCB)}}. IEEE, \bibinfo{publisher}{IEEE}, \bibinfo{address}{USA}, \bibinfo{pages}{474--482}.
\newblock


\bibitem[Bhattacharjee et~al\mbox{.}(2019)]%
        {bhattacharjee2019recent}
\bibfield{author}{\bibinfo{person}{Sushil Bhattacharjee}, \bibinfo{person}{Amir Mohammadi}, \bibinfo{person}{Andr{\'e} Anjos}, {and} \bibinfo{person}{S{\'e}bastien Marcel}.} \bibinfo{year}{2019}\natexlab{}.
\newblock \showarticletitle{Recent Advances in Face Presentation Attack Detection}.
\newblock In \bibinfo{booktitle}{\emph{Handbook of Biometric Anti-Spoofing}}. \bibinfo{publisher}{Springer}, \bibinfo{address}{Germany}, \bibinfo{pages}{207--228}.
\newblock


\bibitem[Bhattacharjee et~al\mbox{.}(2018)]%
        {csmask}
\bibfield{author}{\bibinfo{person}{Sushil Bhattacharjee}, \bibinfo{person}{Amir Mohammadi}, {and} \bibinfo{person}{S{\'{e}}bastien Marcel}.} \bibinfo{year}{2018}\natexlab{}.
\newblock \showarticletitle{Spoofing Deep Face Recognition With Custom Silicone Masks}. In \bibinfo{booktitle}{\emph{{Proceedings of IEEE 9th International Conference on Biometrics: Theory, Applications, and Systems (BTAS)}}}. \bibinfo{publisher}{IEEE}, \bibinfo{address}{{Los Angeles (CA), USA}}, \bibinfo{pages}{1--6}.
\newblock


\bibitem[Boulkenafet et~al\mbox{.}(2018)]%
        {boulkenafet2018generalization}
\bibfield{author}{\bibinfo{person}{Zinelabidine Boulkenafet}, \bibinfo{person}{Jukka Komulainen}, {and} \bibinfo{person}{Abdenour Hadid}.} \bibinfo{year}{2018}\natexlab{}.
\newblock \showarticletitle{On the generalization of color texture-based face anti-spoofing}.
\newblock \bibinfo{journal}{\emph{Image and Vision Computing}}  \bibinfo{volume}{77} (\bibinfo{year}{2018}), \bibinfo{pages}{1--9}.
\newblock


\bibitem[Chen(2005)]%
        {chen2005cashless}
\bibfield{author}{\bibinfo{person}{Grace Chen}.} \bibinfo{year}{2005}\natexlab{}.
\newblock \bibinfo{title}{Cashless payment system}.
\newblock
\newblock
\newblock
\shownote{US Patent App. 10/622,718}.


\bibitem[de~Freitas~Pereira et~al\mbox{.}(2012)]%
        {de2012lbp}
\bibfield{author}{\bibinfo{person}{Tiago de Freitas~Pereira}, \bibinfo{person}{Andr{\'e} Anjos}, \bibinfo{person}{Jos{\'e}~Mario De~Martino}, {and} \bibinfo{person}{S{\'e}bastien Marcel}.} \bibinfo{year}{2012}\natexlab{}.
\newblock \showarticletitle{LBP- TOP based countermeasure against face spoofing attacks}. In \bibinfo{booktitle}{\emph{Asian Conference on Computer Vision}}. Springer, \bibinfo{publisher}{Springer}, \bibinfo{address}{Germany}, \bibinfo{pages}{121--132}.
\newblock


\bibitem[Erdogmus and Marcel(2013a)]%
        {erdogmus2013spoofing}
\bibfield{author}{\bibinfo{person}{Nesli Erdogmus} {and} \bibinfo{person}{S{\'e}bastien Marcel}.} \bibinfo{year}{2013}\natexlab{a}.
\newblock \showarticletitle{Spoofing 2D face recognition systems with 3D masks}. In \bibinfo{booktitle}{\emph{2013 International Conference of the BIOSIG Special Interest Group (BIOSIG)}}. IEEE, \bibinfo{publisher}{IEEE}, \bibinfo{address}{Germany}, \bibinfo{pages}{1--8}.
\newblock


\bibitem[Erdogmus and Marcel(2013b)]%
        {3dmad}
\bibfield{author}{\bibinfo{person}{Nesli Erdogmus} {and} \bibinfo{person}{S{\'e}bastien Marcel}.} \bibinfo{year}{2013}\natexlab{b}.
\newblock \showarticletitle{Spoofing in 2d face recognition with 3d masks and anti-spoofing with kinect}. In \bibinfo{booktitle}{\emph{2013 IEEE Sixth International Conference on Biometrics: Theory, Applications and Systems (BTAS)}}. IEEE, \bibinfo{publisher}{IEEE}, \bibinfo{address}{USA}, \bibinfo{pages}{1--6}.
\newblock


\bibitem[Evans(2019)]%
        {evans2019handbook}
\bibfield{author}{\bibinfo{person}{Nicholas Evans}.} \bibinfo{year}{2019}\natexlab{}.
\newblock \bibinfo{booktitle}{\emph{Handbook of Biometric Anti-Spoofing: Presentation Attack Detection}}.
\newblock \bibinfo{publisher}{Springer}, \bibinfo{address}{Germany}.
\newblock


\bibitem[Fathy et~al\mbox{.}(2015)]%
        {fathy2015face}
\bibfield{author}{\bibinfo{person}{Mohammed~E Fathy}, \bibinfo{person}{Vishal~M Patel}, {and} \bibinfo{person}{Rama Chellappa}.} \bibinfo{year}{2015}\natexlab{}.
\newblock \showarticletitle{Face-based active authentication on mobile devices}. In \bibinfo{booktitle}{\emph{2015 IEEE International Conference on Acoustics, Speech and Signal Processing (ICASSP)}}. IEEE, \bibinfo{publisher}{IEEE}, \bibinfo{address}{Australia}, \bibinfo{pages}{1687--1691}.
\newblock


\bibitem[George and Marcel(2019)]%
        {deep2}
\bibfield{author}{\bibinfo{person}{Anjith George} {and} \bibinfo{person}{S{\'e}bastien Marcel}.} \bibinfo{year}{2019}\natexlab{}.
\newblock \showarticletitle{Deep pixel-wise binary supervision for face presentation attack detection}. In \bibinfo{booktitle}{\emph{2019 international conference on biometrics (ICB)}}. IEEE, \bibinfo{publisher}{IEEE}, \bibinfo{address}{Greece}, \bibinfo{pages}{1--8}.
\newblock


\bibitem[Hernandez-Ortega et~al\mbox{.}(2019)]%
        {hernandez2019introduction}
\bibfield{author}{\bibinfo{person}{Javier Hernandez-Ortega}, \bibinfo{person}{Julian Fierrez}, \bibinfo{person}{Aythami Morales}, {and} \bibinfo{person}{Javier Galbally}.} \bibinfo{year}{2019}\natexlab{}.
\newblock \showarticletitle{Introduction to face presentation attack detection}.
\newblock In \bibinfo{booktitle}{\emph{Handbook of Biometric Anti-Spoofing}}. \bibinfo{publisher}{Springer}, \bibinfo{address}{Germany}, \bibinfo{pages}{187--206}.
\newblock


\bibitem[Jain and Li(2011)]%
        {jain2011handbook}
\bibfield{author}{\bibinfo{person}{Anil~K Jain} {and} \bibinfo{person}{Stan~Z Li}.} \bibinfo{year}{2011}\natexlab{}.
\newblock \bibinfo{booktitle}{\emph{Handbook of face recognition}}.
\newblock \bibinfo{publisher}{Springer}, \bibinfo{address}{Germany}.
\newblock


\bibitem[Jain and Nandakumar(2012)]%
        {jain2012biometric}
\bibfield{author}{\bibinfo{person}{Anil~K Jain} {and} \bibinfo{person}{Karthik Nandakumar}.} \bibinfo{year}{2012}\natexlab{}.
\newblock \showarticletitle{Biometric Authentication: System Security and User Privacy.}
\newblock \bibinfo{journal}{\emph{IEEE Computer}} \bibinfo{volume}{45}, \bibinfo{number}{11} (\bibinfo{year}{2012}), \bibinfo{pages}{87--92}.
\newblock


\bibitem[Jain et~al\mbox{.}(2006)]%
        {jain2006biometrics}
\bibfield{author}{\bibinfo{person}{Anil~K Jain}, \bibinfo{person}{Arun Ross}, {and} \bibinfo{person}{Sharath Pankanti}.} \bibinfo{year}{2006}\natexlab{}.
\newblock \showarticletitle{Biometrics: a tool for information security}.
\newblock \bibinfo{journal}{\emph{IEEE transactions on information forensics and security}} \bibinfo{volume}{1}, \bibinfo{number}{2} (\bibinfo{year}{2006}), \bibinfo{pages}{125--143}.
\newblock


\bibitem[Jain et~al\mbox{.}(2004)]%
        {jain2004introduction}
\bibfield{author}{\bibinfo{person}{Anil~K Jain}, \bibinfo{person}{Arun Ross}, \bibinfo{person}{Salil Prabhakar}, {et~al\mbox{.}}} \bibinfo{year}{2004}\natexlab{}.
\newblock \showarticletitle{An introduction to biometric recognition}.
\newblock \bibinfo{journal}{\emph{IEEE Transactions on circuits and systems for video technology}} \bibinfo{volume}{14}, \bibinfo{number}{1} (\bibinfo{year}{2004}), \bibinfo{pages}{12--20}.
\newblock


\bibitem[Komulainen et~al\mbox{.}(2019)]%
        {komulainen2019review}
\bibfield{author}{\bibinfo{person}{Jukka Komulainen}, \bibinfo{person}{Zinelabidine Boulkenafet}, {and} \bibinfo{person}{Zahid Akhtar}.} \bibinfo{year}{2019}\natexlab{}.
\newblock \showarticletitle{Review of face presentation attack detection competitions}.
\newblock In \bibinfo{booktitle}{\emph{Handbook of Biometric Anti-Spoofing}}. \bibinfo{publisher}{Springer}, \bibinfo{address}{Germany}, \bibinfo{pages}{291--317}.
\newblock


\bibitem[Kotwal et~al\mbox{.}(2019)]%
        {kotwal2019multispectral}
\bibfield{author}{\bibinfo{person}{Ketan Kotwal}, \bibinfo{person}{Sushil Bhattacharjee}, {and} \bibinfo{person}{S{\'e}bastien Marcel}.} \bibinfo{year}{2019}\natexlab{}.
\newblock \showarticletitle{Multispectral Deep Embeddings as a Countermeasure to Custom Silicone Mask Presentation Attacks}.
\newblock \bibinfo{journal}{\emph{IEEE Transactions on Biometrics, Behavior, and Identity Science}} \bibinfo{volume}{1}, \bibinfo{number}{4} (\bibinfo{year}{2019}), \bibinfo{pages}{238--251}.
\newblock


\bibitem[Li et~al\mbox{.}(2018)]%
        {rose}
\bibfield{author}{\bibinfo{person}{Haoliang Li}, \bibinfo{person}{Wen Li}, \bibinfo{person}{Hong Cao}, \bibinfo{person}{Shiqi Wang}, \bibinfo{person}{Feiyue Huang}, {and} \bibinfo{person}{Alex~C Kot}.} \bibinfo{year}{2018}\natexlab{}.
\newblock \showarticletitle{Unsupervised domain adaptation for face anti-spoofing}.
\newblock \bibinfo{journal}{\emph{IEEE Transactions on Information Forensics and Security}} \bibinfo{volume}{1}, \bibinfo{number}{15} (\bibinfo{year}{2018}), \bibinfo{pages}{1794--1809}.
\newblock


\bibitem[Li et~al\mbox{.}(2016)]%
        {S1}
\bibfield{author}{\bibinfo{person}{Lei Li}, \bibinfo{person}{Xiaoyi Feng}, \bibinfo{person}{Zinelabidine Boulkenafet}, \bibinfo{person}{Zhaoqiang Xia}, \bibinfo{person}{Mingming Li}, {and} \bibinfo{person}{Abdenour Hadid}.} \bibinfo{year}{2016}\natexlab{}.
\newblock \showarticletitle{An original face anti-spoofing approach using partial convolutional neural network}. In \bibinfo{booktitle}{\emph{2016 Sixth International Conference on Image Processing Theory, Tools and Applications (IPTA)}}. IEEE, \bibinfo{publisher}{IEEE}, \bibinfo{address}{Finland}, \bibinfo{pages}{1--6}.
\newblock


\bibitem[Liu and Kumar(2018)]%
        {liu2018detecting}
\bibfield{author}{\bibinfo{person}{Jun Liu} {and} \bibinfo{person}{Ajay Kumar}.} \bibinfo{year}{2018}\natexlab{}.
\newblock \showarticletitle{Detecting presentation attacks from 3d face masks under multispectral imaging}. In \bibinfo{booktitle}{\emph{Proceedings of the IEEE Conference on Computer Vision and Pattern Recognition Workshops}}. \bibinfo{publisher}{IEEE}, \bibinfo{address}{USA}, \bibinfo{pages}{47--52}.
\newblock


\bibitem[Liu et~al\mbox{.}(2018)]%
        {rppg}
\bibfield{author}{\bibinfo{person}{Yaojie Liu}, \bibinfo{person}{Amin Jourabloo}, {and} \bibinfo{person}{Xiaoming Liu}.} \bibinfo{year}{2018}\natexlab{}.
\newblock \showarticletitle{Learning deep models for face anti-spoofing: Binary or auxiliary supervision}. In \bibinfo{booktitle}{\emph{Proceedings of the IEEE Conference on Computer Vision and Pattern Recognition}}. \bibinfo{publisher}{IEEE}, \bibinfo{address}{USA}, \bibinfo{pages}{389--398}.
\newblock


\bibitem[Mahbub et~al\mbox{.}(2016)]%
        {mahbub2016active}
\bibfield{author}{\bibinfo{person}{Upal Mahbub}, \bibinfo{person}{Sayantan Sarkar}, \bibinfo{person}{Vishal~M Patel}, {and} \bibinfo{person}{Rama Chellappa}.} \bibinfo{year}{2016}\natexlab{}.
\newblock \showarticletitle{Active user authentication for smartphones: A challenge data set and benchmark results}. In \bibinfo{booktitle}{\emph{2016 IEEE 8th International Conference on Biometrics Theory, Applications and Systems (BTAS)}}. IEEE, \bibinfo{publisher}{IEEE}, \bibinfo{address}{USA}, \bibinfo{pages}{1--8}.
\newblock


\bibitem[Majumdar et~al\mbox{.}(2019)]%
        {majumdar2019evading}
\bibfield{author}{\bibinfo{person}{Puspita Majumdar}, \bibinfo{person}{Akshay Agarwal}, \bibinfo{person}{Richa Singh}, {and} \bibinfo{person}{Mayank Vatsa}.} \bibinfo{year}{2019}\natexlab{}.
\newblock \showarticletitle{Evading Face Recognition via Partial Tampering of Faces}. In \bibinfo{booktitle}{\emph{Proceedings of the IEEE Conference on Computer Vision and Pattern Recognition Workshops}}. \bibinfo{publisher}{IEEE}, \bibinfo{address}{USA}, \bibinfo{pages}{0--0}.
\newblock


\bibitem[{Manjani} et~al\mbox{.}(2017)]%
        {ddgl}
\bibfield{author}{\bibinfo{person}{I. {Manjani}}, \bibinfo{person}{S. {Tariyal}}, \bibinfo{person}{M. {Vatsa}}, \bibinfo{person}{R. {Singh}}, {and} \bibinfo{person}{A. {Majumdar}}.} \bibinfo{year}{2017}\natexlab{}.
\newblock \showarticletitle{Detecting Silicone Mask-Based Presentation Attack via Deep Dictionary Learning}.
\newblock \bibinfo{journal}{\emph{IEEE Transactions on Information Forensics and Security}}  \bibinfo{volume}{12} (\bibinfo{date}{July} \bibinfo{year}{2017}), \bibinfo{pages}{1713--1723}.
\newblock
\showISSN{1556-6013}


\bibitem[Marcel et~al\mbox{.}(2014)]%
        {marcel2014handbook}
\bibfield{author}{\bibinfo{person}{S{\'e}bastien Marcel}, \bibinfo{person}{Mark~S Nixon}, {and} \bibinfo{person}{Stan~Z Li}.} \bibinfo{year}{2014}\natexlab{}.
\newblock \bibinfo{booktitle}{\emph{Handbook of Biometric Anti-Spoofing}}.
\newblock \bibinfo{publisher}{Springer}, \bibinfo{address}{Germany}.
\newblock


\bibitem[Mehta et~al\mbox{.}(2019)]%
        {mehta2019crafting}
\bibfield{author}{\bibinfo{person}{Suril Mehta}, \bibinfo{person}{Anannya Uberoi}, \bibinfo{person}{Akshay Agarwal}, \bibinfo{person}{Mayank Vatsa}, {and} \bibinfo{person}{Richa Singh}.} \bibinfo{year}{2019}\natexlab{}.
\newblock \showarticletitle{Crafting a panoptic face presentation attack detector}.
\newblock \bibinfo{journal}{\emph{IEEE}} \bibinfo{volume}{1}, \bibinfo{number}{5} (\bibinfo{year}{2019}), \bibinfo{pages}{1--6}.
\newblock


\bibitem[Ramachandra and Busch(2017)]%
        {ramachandra2017presentation}
\bibfield{author}{\bibinfo{person}{Raghavendra Ramachandra} {and} \bibinfo{person}{Christoph Busch}.} \bibinfo{year}{2017}\natexlab{}.
\newblock \showarticletitle{Presentation attack detection methods for face recognition systems: A comprehensive survey}.
\newblock \bibinfo{journal}{\emph{ACM Computing Surveys (CSUR)}} \bibinfo{volume}{50}, \bibinfo{number}{1} (\bibinfo{year}{2017}), \bibinfo{pages}{8}.
\newblock


\bibitem[Ramachandra et~al\mbox{.}(2019)]%
        {psan}
\bibfield{author}{\bibinfo{person}{Raghavendra Ramachandra}, \bibinfo{person}{Sushma Venkatesh}, \bibinfo{person}{Kiran~B Raja}, \bibinfo{person}{Sushil Bhattacharjee}, \bibinfo{person}{Pankaj Wasnik}, \bibinfo{person}{Sebastien Marcel}, {and} \bibinfo{person}{Christoph Busch}.} \bibinfo{year}{2019}\natexlab{}.
\newblock \showarticletitle{Custom silicone Face Masks: Vulnerability of Commercial Face Recognition Systems \& Presentation Attack Detection}. In \bibinfo{booktitle}{\emph{2019 7th International Workshop on Biometrics and Forensics (IWBF)}}. IEEE, \bibinfo{publisher}{IEEE}, \bibinfo{address}{USA}, \bibinfo{pages}{1--6}.
\newblock


\bibitem[Rathgeb et~al\mbox{.}(2019)]%
        {rathgeb2019impact}
\bibfield{author}{\bibinfo{person}{Christian Rathgeb}, \bibinfo{person}{Antitza Dantcheva}, {and} \bibinfo{person}{Christoph Busch}.} \bibinfo{year}{2019}\natexlab{}.
\newblock \showarticletitle{Impact and Detection of Facial Beautification in Face Recognition: An Overview}.
\newblock \bibinfo{journal}{\emph{IEEE Access}}  \bibinfo{volume}{7} (\bibinfo{year}{2019}), \bibinfo{pages}{152667--152678}.
\newblock


\bibitem[Rehman et~al\mbox{.}(2018)]%
        {livenet}
\bibfield{author}{\bibinfo{person}{Yasar Abbas~Ur Rehman}, \bibinfo{person}{Lai~Man Po}, {and} \bibinfo{person}{Mengyang Liu}.} \bibinfo{year}{2018}\natexlab{}.
\newblock \showarticletitle{LiveNet: Improving features generalization for face liveness detection using convolution neural networks}.
\newblock \bibinfo{journal}{\emph{Expert Systems with Applications}} \bibinfo{volume}{10}, \bibinfo{number}{5} (\bibinfo{year}{2018}), \bibinfo{pages}{159--169}.
\newblock


\bibitem[Vareto et~al\mbox{.}(2019)]%
        {vareto2019swax}
\bibfield{author}{\bibinfo{person}{Rafael~Henrique Vareto}, \bibinfo{person}{Araceli~Marcia Sandanha}, {and} \bibinfo{person}{William~Robson Schwartz}.} \bibinfo{year}{2019}\natexlab{}.
\newblock \showarticletitle{The SWAX Benchmark: Attacking Biometric Systems with Wax Figures}.
\newblock \bibinfo{journal}{\emph{arXiv preprint arXiv:1910.09642}}  \bibinfo{volume}{6} (\bibinfo{year}{2019}), \bibinfo{pages}{1222--1232}.
\newblock


\bibitem[Xiang and Zhu(2017)]%
        {mtcnn}
\bibfield{author}{\bibinfo{person}{Jia Xiang} {and} \bibinfo{person}{Gengming Zhu}.} \bibinfo{year}{2017}\natexlab{}.
\newblock \showarticletitle{Joint Face Detection and Facial Expression Recognition with MTCNN}. In \bibinfo{booktitle}{\emph{2017 4th International Conference on Information Science and Control Engineering (ICISCE)}}. IEEE, \bibinfo{publisher}{IEEE}, \bibinfo{address}{China}, \bibinfo{pages}{424--427}.
\newblock


\end{thebibliography}

\end{document}